\documentclass{article}
\PassOptionsToPackage{hyphens}{url}
\usepackage{microtype,graphicx,subfigure,booktabs,hyperref,amsmath,amssymb,mathtools,bm}
\usepackage[capitalize,noabbrev]{cleveref}

\usepackage[accepted]{icml2025}
\icmltitlerunning{Scalable-Softmax Is Superior for Attention}
\begin{document}
\twocolumn[
\icmltitle{Scalable-Softmax Is Superior for Attention}
\begin{icmlauthorlist}
\icmlauthor{Ken M. Nakanishi}{ipi}
\end{icmlauthorlist}
\icmlaffiliation{ipi}{Institute for Physics of Intelligence, The University of Tokyo, Tokyo 113-0033, Japan}
\icmlcorrespondingauthor{Ken M. Nakanishi}{ken-nakanishi@g.ecc.u-tokyo.ac.jp}
\icmlkeywords{Softmax, Length generalization, Attention score, Attention, Long context, Needle-In-A-Haystack}
\vskip 0.3in
]
\makeatletter\renewcommand{\Notice@String}{}\makeatother
\printAffiliationsAndNotice{}

\begin{abstract}
    The maximum element of the vector output by the Softmax function approaches zero as the input vector size increases. Transformer-based language models rely on Softmax to compute attention scores, causing the attention distribution to flatten as the context size grows. This reduces the model's ability to prioritize key information effectively and potentially limits its length generalization. To address this problem, we propose Scalable-Softmax (SSMax), which replaces Softmax in scenarios where the input vector size varies. SSMax can be seamlessly integrated into existing Transformer-based architectures. Experimental results in language modeling show that models using SSMax not only achieve faster loss reduction during pretraining but also significantly improve performance in long contexts and key information retrieval. Furthermore, an analysis of attention scores reveals that SSMax enables the model to focus attention on key information even in long contexts. Additionally, although models that use SSMax from the beginning of pretraining achieve better length generalization, those that have already started pretraining can still gain some of this ability by replacing Softmax in the attention layers with SSMax, either during or after pretraining.
\end{abstract}

\section{Introduction}

\begin{figure}[t]
    \centering
    \includegraphics[width=\columnwidth]{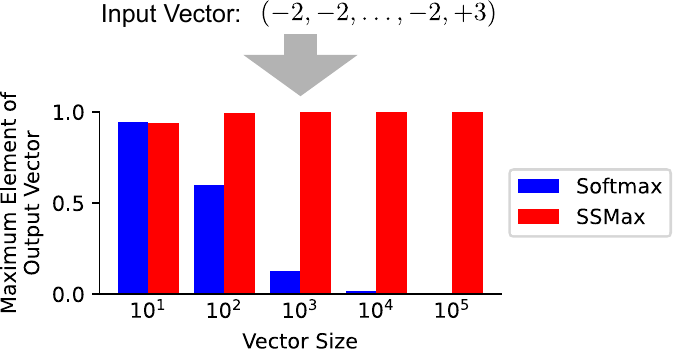}
    \vspace*{-2ex}
    \caption{
        Comparison of Softmax and SSMax, illustrating the issue of attention fading and the effectiveness of SSMax in preventing it. As the input vector size increases, the maximum value of the output vector produced by Softmax decreases, demonstrating the problem of attention fading. In contrast, SSMax keeps the maximum value close to 1, regardless of the input size. The input vector consists of -2 for all elements except the last, which is set to +3. The scaling parameter $s$ of SSMax is set to 0.43.
    }
    \label{fig:key}
\end{figure}

Length generalization, the ability to handle longer context sizes than those used during training, is a key challenge for Transformer-based large language models (LLMs)~\cite{vaswani2017attention,kazemnejad2023impact}. Processing longer contexts enhances LLMs' ability to leverage in-context learning~\cite{brown2020language} and chain-of-thought reasoning~\cite{wei2022chain}. However, the computational and memory requirements for training Transformers grow quadratically with context size, imposing practical limits on the context sizes used during training. Consequently, LLMs must acquire the ability to generalize to longer context sizes beyond those seen during training. There are four primary approaches to addressing length generalization: improving positional encoding methods~\cite{wei2022chain,press2021train,su2024roformer,kazemnejad2023impact,shaw2018self}, adopting sparse attention mechanisms~\cite{ainslie2020etc,gupta2020gmat,zaheer2020big,beltagy2020longformer,child2019generating,kitaev2020reformer,roy2021efficient,sukhbaatar2019adaptive}, further training on longer contexts after modifying positional encodings~\cite{liu2024scaling,liu2024world,ye2024differential}, and enhancing attention mechanisms~\cite{ye2024differential,wang2024length}. This work focuses on enhancing attention mechanisms as an approach to address length generalization, specifically by replacing the Softmax function within the attention layers of Transformers.

Softmax converts an input vector into a vector that can be interpreted as a probability distribution, where all elements are non-negative and sum to one. In deep learning, Softmax is commonly used in multi-class classification tasks to convert logits into a probability distribution~\cite{lecun1998gradient}. In the attention layers of Transformer-based language models, Softmax computes a probability distribution over all tokens in the context, determining how much attention is allocated to each token~\cite{vaswani2017attention}. Unlike the Softmax used in classification tasks, where the input vector size is fixed, the Softmax in attention layers has an input vector size that varies with the context size. The denominator of Softmax is the sum of the exponentials of all input elements, and its value increases as the context size grows, due to the larger number of input elements. In contrast, the numerator, the exponential of a single input element, is independent of the input vector size. As the input vector size grows, the resulting probability distribution becomes increasingly flat. In this paper, we refer to this phenomenon as \textit{attention fading}, which we hypothesize reduces the model's ability to focus on key tokens in the context, thereby limiting length generalization.

To address this issue, we propose Scalable-Softmax (SSMax), a novel alternative to Softmax in attention layers that mitigates attention fading and enhances length generalization. SSMax transforms an input vector into a probability distribution, similar to Softmax, but avoids attention fading even for large input vector sizes. \Cref{fig:key} illustrates the effectiveness of SSMax, contrasting it with Softmax, which suffers from attention fading. The name `Scalable-Softmax' reflects its ability to handle varying input vector sizes while preventing attention fading. Note that `Scalable' does not refer to the scaling parameter $s$, which will be introduced later. Additionally, as shown in \cref{sec:impl}, SSMax integrates seamlessly into existing architectures with minimal code modifications, maintaining compatibility and efficiency.

In this study, we conducted several evaluations to assess the effectiveness of SSMax in improving Transformer performance. First, we compared the learning curves of Transformer models during pretraining and found that those with SSMax consistently achieved lower loss values compared to the standard Transformer. In all subsequent evaluations, we modified RoPE's $\theta$ to 50 times its training value without additional training, allowing us to assess the model's ability to generalize to large context sizes under this drastic change. Second, unlike the standard Transformer, models with SSMax maintained significantly lower loss values even as the context size increased well beyond the training sequence length. Third, in key information retrieval tasks, models with SSMax retrieved key information reasonably accurately, even when the context size was extended up to 10 times the training sequence length. Further analysis of attention scores revealed that models with SSMax allocate significantly more attention to key tokens, even in long-context scenarios. Lastly, while models trained with SSMax from the beginning achieve superior length generalization, we showed that replacing Softmax with SSMax in pretrained models can still provide some degree of improvement.

\section{Scalable-Softmax (SSMax)}\label{sec:ssmax}

The Softmax function transforms an input vector into a vector that can be interpreted as a probability distribution, where all elements are non-negative and sum to one. For an input vector $\bm{z}$ of size $n$, with components $z_i\ (i=1,2,\dots,n)$, Softmax is defined as follows:
\begin{equation}\label{eq:softmax}
    z_i \mapsto \frac{e^{z_i}}{\sum_{j=1}^n e^{z_j}}.
\end{equation}
In the attention layers of Transformers, the input vector size $n$ increases as the context size grows. Softmax plays a critical role in computing a probability distribution over all tokens in the context, determining how much attention is allocated to each token. When $n$ grows, the denominator in \cref{eq:softmax} increases, while the numerator remains independent of $n$. As a result, the resulting probability distribution becomes increasingly flat. This phenomenon, which we refer to as \textit{attention fading}, reduces the model's ability to focus on key tokens within the context, potentially limiting its length generalization capability.

To address this issue, we propose the Scalable-Softmax (SSMax) function, defined as:
\begin{equation}\label{eq:ssmax}
z_i \mapsto \frac{n^{sz_i}}{\sum_{j=1}^n n^{sz_j}} = \frac{e^{(s\log n)z_i}}{\sum_{j=1}^n e^{(s\log n)z_j}},
\end{equation}
where $s$ is a scalar referred to as the scaling parameter. Similar to Softmax, SSMax transforms an input vector into a probability distribution. However, the key difference lies in the dependence of the exponential base on input vector size $n$. This design aims to prevent attention fading by incorporating the input vector size into the function's formulation, which helps balance the scaling between the numerator and denominator in \cref{eq:softmax}. As a result, the probability distribution remains focused on key tokens even as the input size grows.

In the following subsections, we provide experimental evidence to justify the design of SSMax (\cref{sec:explain_experimental}) and present a theoretical analysis to further explain its design and effectiveness in addressing attention fading (\cref{sec:explain_theoretical}). Finally, we discuss its ease of implementation, requiring minimal modifications to existing architectures (\cref{sec:impl}).

\subsection{Rationale Behind the Design of SSMax}\label{sec:explain_experimental}

\begin{figure}[t]
    \includegraphics[width=0.95\columnwidth]{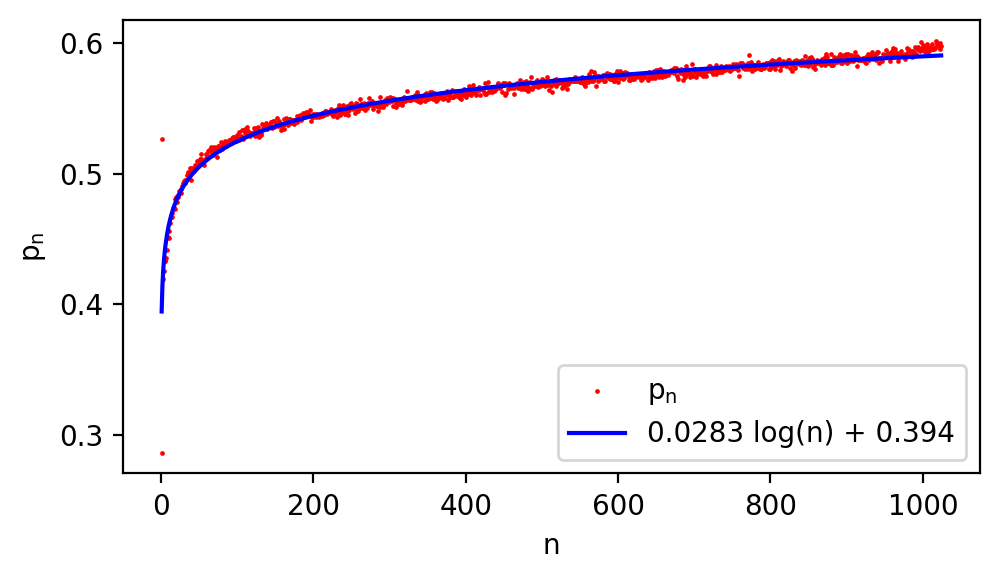}
    \vspace*{-2ex}
    \caption{
        Relationship between $p_n$ and the input vector size $n$. The red dots represent the learned values of $p_n$ after training, and the blue curve is a fitted logarithmic function of the form $p_n \approx a_1 \log n + a_2$. This result suggests that $p_n$ depends logarithmically on $n$, motivating the reformulation of Softmax in \cref{eq:ssmax_with_bias}.
    }
    \label{fig:fit}
\end{figure}

To investigate the optimal formulation of SSMax, we conducted experiments to analyze how attention scores should ideally depend on the input vector size $n$. Specifically, we replaced Softmax in all attention layers with the following modified formulation:
\begin{equation}\label{eq:softmax_with_pn}
    z_i \mapsto \frac{e^{(sp_n+b)z_i}}{\sum_{j=1}^n e^{(sp_n+b)z_j}},
\end{equation}
where $s$ and $b$ are scalar learnable parameters unique to each layer and head, and $p_n\ (n=1,2,\dots,N)$ represents learnable parameters shared across all layers and heads, depending solely on the input vector size $n$. Here, $N$ denotes the sequence length used during training. Since sparse attention mechanisms were not employed in this study, $n$ corresponds to the context size.

We initialized all $p_n$ to 1, $s$ to 1, and $b$ to 0. The model has 162M parameters, and its architecture details are described in \cref{apx:model_param}. Training was conducted with a sequence length of $N=1024$ and a batch size of 2048 for 25,000 iterations (approximately 52B tokens). Detailed training hyperparameters are provided in \cref{apx:pt_param}. After training, we plotted the values of $p_n$, as shown in \cref{fig:fit}.

\Cref{fig:fit} demonstrates that $p_n$ closely follows a logarithmic relationship of the form $p_n \approx a_1 \log n + a_2$, where $a_1$ and $a_2$ are fitted constants. This finding suggests that the attention mechanism in Transformers could benefit from reformulating Softmax as:
\begin{equation}\label{eq:ssmax_with_bias}
    z_i \mapsto \frac{e^{(s\log n+b)z_i}}{\sum_{j=1}^n e^{(s\log n+b)z_j}} = \frac{n^{sz_i}e^{bz_i}}{\sum_{j=1}^n n^{sz_j}e^{bz_j}},
\end{equation}
where $s$ and $b$ are layer- and head-specific learnable scalar parameters. We refer to $b$ as the bias parameter in this formulation.

Based on these findings, further evaluations conducted in \cref{sec:evaluations} reveal that while the inclusion of the bias parameter $b$ slightly accelerates loss reduction during pretraining, omitting $b$ leads to better length generalization performance.

\subsection{Justification for the Design of SSMax}\label{sec:explain_theoretical}

\begin{figure}[t]
    \centering
    \includegraphics[width=\columnwidth]{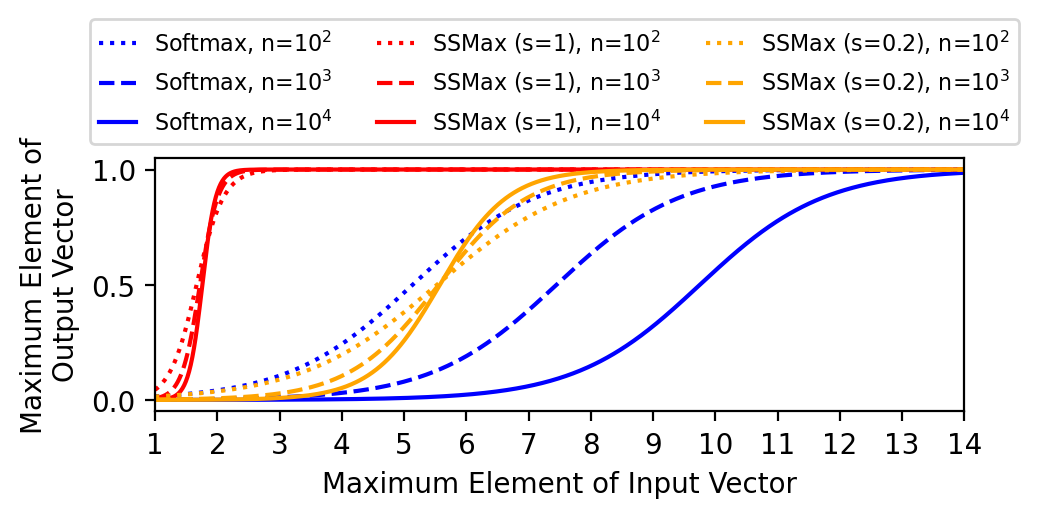}
    \vspace*{-2ex}
    \caption{
        An example illustrating the behavior of Softmax and SSMax for an input vector of size $n$ given by $(0, \frac{1}{n-2}, \frac{2}{n-2}, \dots, \frac{n-1}{n-2}, 1, z_\mathrm{max})$. The horizontal axis represents the value of $z_\mathrm{max}$, while the vertical axis represents its transformed value. The red and orange lines correspond to SSMax with different scaling parameters $s$, and the blue lines correspond to Softmax, with line styles indicating different input vector sizes. This figure demonstrates that, under Softmax, the value of $z_\mathrm{max}$ required to focus attention increases indefinitely as $n$ grows. In contrast, SSMax ensures that attention is focused as long as $z_\mathrm{max}$ exceeds the other values by approximately $\frac{1}{s}$, regardless of $n$.
    }
    \label{fig:concept}
\end{figure}

Let $\bm{z} = (z_1, z_2, \dots, z_n)$ be an input vector of size $n$, where $z_\mathrm{max}$, $z_\mathrm{2nd}$, and $z_\mathrm{min}$ denote its maximum, second largest, and minimum elements, respectively. For simplicity, we assume $z_\mathrm{max} > z_\mathrm{2nd}$ in the following analysis. When $\bm{z}$ is processed by Softmax, $z_\mathrm{max}$ is transformed as
\begin{equation}\label{eq:explain_softmax}
    z_\mathrm{max} \mapsto \frac{e^{z_\mathrm{max}}}{\sum_{j=1}^n e^{z_j}}.
\end{equation}
The right-hand side of \cref{eq:explain_softmax} can be evaluated as
\begin{align}
    \frac{e^{z_\mathrm{max}}}{\sum_{j=1}^n e^{z_j}}
    &\leq \frac{e^{z_\mathrm{max}}}{(n-1)e^{z_\mathrm{min}} + e^{z_\mathrm{max}}} \notag \\
    &= \frac{1}{\frac{n-1}{e^{z_\mathrm{max} - z_\mathrm{min}}} + 1}. \label{eq:explain_softmax_approx}
\end{align}
From \cref{eq:explain_softmax_approx}, it follows that the maximum element of the output vector produced by Softmax approaches zero as the input vector size $n$ increases.

On the other hand, when $\bm{z}$ is processed by SSMax, $z_\mathrm{max}$ is transformed as
\begin{equation}\label{eq:explain_ssmax}
    z_\mathrm{max} \mapsto \frac{n^{sz_\mathrm{max}}}{\sum_{j=1}^n n^{sz_j}}.
\end{equation}
Assuming $s > 0$ for simplicity\footnote{For $s < 0$, a similar argument applies by substituting $s \mapsto -s$ and $\bm{z} \mapsto -\bm{z}$.}, the right-hand side of \cref{eq:explain_ssmax} can be evaluated as
\begin{align}
    \frac{n^{sz_\mathrm{max}}}{\sum_{j=1}^n n^{sz_j}}
    &\leq \frac{n^{sz_\mathrm{max}}}{(n-1)n^{sz_\mathrm{min}} + n^{sz_\mathrm{max}}} \notag \\
    &= \frac{1}{\frac{n-1}{n^{s(z_\mathrm{max} - z_\mathrm{min})}} + 1}, \label{eq:explain_ssmax_approx_1}
\end{align}
\begin{align}
    \frac{n^{sz_\mathrm{max}}}{\sum_{j=1}^n n^{sz_j}}
    &\geq \frac{n^{sz_\mathrm{max}}}{(n-1)n^{sz_\mathrm{2nd}} + n^{sz_\mathrm{max}}} \notag \\
    &= \frac{1}{\frac{n-1}{n^{s(z_\mathrm{max} - z_\mathrm{2nd})}} + 1}. \label{eq:explain_ssmax_approx_2}
\end{align}
From \cref{eq:explain_ssmax_approx_1,eq:explain_ssmax_approx_2}, the maximum element of the output vector produced by SSMax exhibits the following properties as $n$ increases:
\begin{itemize}
    \item If $z_\mathrm{max} - z_\mathrm{2nd} > \frac{1}{s}$, the maximum value approaches 1, indicating that attention is focused on the element with the highest value.
    \item If $z_\mathrm{max} - z_\mathrm{min} < \frac{1}{s}$, the maximum value approaches 0, meaning that attention is distributed across all elements.
\end{itemize}
Thus, SSMax ensures that attention is focused on elements whose values exceed others by approximately $\frac{1}{s}$, while distributing attention when all values are within a range of approximately $\frac{1}{s}$. This design allows the model to adapt its attention allocation dynamically, focusing on key tokens when significant differences exist or distributing attention when the input values are relatively uniform.

For instance, in \cref{fig:key}, since the condition $z_\mathrm{max} - z_\mathrm{2nd} > \frac{1}{s}$ is satisfied, the maximum element of the output vector produced by SSMax approaches 1 as the input vector size $n$ increases. Another example is illustrated in \cref{fig:concept}, where the input vector of size $n$ is given by $(0, \frac{1}{n-2}, \frac{2}{n-2}, \dots, \frac{n-1}{n-2}, 1, z_\mathrm{max})$. The figure demonstrates that, under Softmax, the value of $z_\mathrm{max}$ required to focus attention on the corresponding element grows indefinitely as $n$ increases. In contrast, SSMax ensures that attention is focused on the element associated with $z_\mathrm{max}$ as long as it exceeds other values by approximately $\frac{1}{s}$, regardless of $n$. This property allows SSMax to effectively allocate attention to key elements, even as the input vector size increases significantly.

\subsection{Seamless Implementation of SSMax}\label{sec:impl}

The implementation of SSMax is straightforward and requires minimal modifications to existing Transformer architectures. For simplicity, we describe the standard dense attention mechanism without sparse attention variants.

In standard Transformers, the attention scores for the $n$-th token $\bm{a}_n \in \mathbb{R}^{n}$ are typically computed using the query vector of the $n$-th token $\bm{q}_n \in \mathbb{R}^{d}$ and the key tensor $K_{1:n} \in \mathbb{R}^{n \times d}$, which contains the key vectors of the first $n$ tokens, as
\begin{equation}\label{eq:softmax_impl}
\bm{a}_n = \mathrm{Softmax}\left( \frac{\bm{q}_n K_{1:n}^T}{\sqrt{d}} \right).
\end{equation}
When replacing Softmax with SSMax, the attention computation can be reformulated based on \cref{eq:ssmax} as
\begin{align}
\bm{a}_n &= \mathrm{SSMax}\left( \frac{\bm{q}_n K_{1:n}^T}{\sqrt{d}} \right) \notag \\
&= \mathrm{Softmax}\left( \frac{(s \log n) \bm{q}_n K_{1:n}^T}{\sqrt{d}} \right), \label{eq:ssmax_impl}
\end{align}
where $s \in \mathbb{R}$ is a learnable scaling parameter.

As shown in \cref{eq:ssmax_impl}, replacing Softmax with SSMax simply requires multiplying the query vector $\bm{q}_n$ by $s \log n$. This simplicity ensures that SSMax can be seamlessly integrated into existing architectures with minimal modifications.

\section{Evaluations}\label{sec:evaluations}

To evaluate the impact of replacing Softmax with SSMax in the attention layers, we conducted a series of experiments focusing on long-context generalization and attention mechanisms. First, we compared the learning curves during pretraining to assess whether SSMax improves training efficiency (\cref{sec:learning_curve}). Next, we analyzed the model's ability to generalize to longer sequences by measuring per-position test loss on sequences that were approximately 20 times longer than during training (\cref{sec:posloss}). We then evaluated key information retrieval performance using the Needle-In-A-Haystack test, assessing whether SSMax improves key information extraction even when the context size is extended up to approximately 10 times the training sequence length (\cref{sec:niah}). Finally, we analyzed attention score allocation to assess how SSMax influences focus on key information during retrieval tasks (\cref{sec:needle_score}).

\paragraph{Transformer Architecture and Configurations}
The Transformer architecture used in these experiments includes enhancements similar to those in Llama 2~\cite{touvron2023llama}, such as RoPE~\cite{su2024roformer}, RMSNorm~\cite{zhang2019root}, SwiGLU~\cite{shazeer2020glu,ramachandran2017searching}, and the removal of bias in projection layers. The model has 12 layers, 12 attention heads, and 162M parameters. RoPE was initialized with $\theta = 10,000$. Details of the model configuration are provided in \cref{apx:model_param}. For tokenization, we employed the GPT-2 tokenizer~\cite{radford2019language}, and the training sequence length was set to 1024.

\paragraph{Evaluated Configurations}
We evaluated the following six variants, each replacing Softmax in the attention layers:
\begin{itemize}
    \item[(a)] \textbf{Softmax}:
    The standard Softmax function, as defined in \cref{eq:softmax}.

    \item[(b)] \textbf{SSMax}:
    The proposed Scalable-Softmax (SSMax) function, as defined in \cref{eq:ssmax}. The scaling parameter $s$ is modeled as a learnable scalar, independently learned for each attention layer and head. Since the model consists of 12 layers and 12 heads, this modification adds only 144 additional parameters, which is negligible compared to the total model size of 162M.

    \item[(c)] \textbf{SSMax without Scaling Parameter}:
    A simplified SSMax variant with the scaling parameter $s$ removed, equivalent to substituting $s = 1$ in \cref{eq:ssmax}.

    \item[(d)] \textbf{SSMax with Bias Parameter}:
    A variant of SSMax that incorporates a learnable bias parameter $b$, as defined in \cref{eq:ssmax_with_bias}. Both $s$ and $b$ are learnable scalars, each specific to every layer and head, resulting in an additional 288 parameters in total.

    \item[(e)] \textbf{Softmax Replaced by SSMax After Pretraining}:
    After pretraining, Softmax is replaced with SSMax. The scaling parameter $s$ is initialized to the reciprocal of the average $\log n$ value during training, which is approximately 0.168, as given by $\frac{1024}{\sum_{n=1}^{1024} \log n} \simeq 0.168$.

    \item[(f)] \textbf{Softmax Replaced by SSMax During Pretraining}:
    The model is first pretrained with Softmax for 175,000 iterations, then switched to SSMax and further trained for 25,000 iterations. The scaling parameter $s$ is initialized as in (e), and a 1000-iteration learning rate warmup is applied following the switch.
\end{itemize}

\subsection{Learning Curve Analysis}\label{sec:learning_curve}

\begin{figure}[t]
    \includegraphics[width=0.95\columnwidth]{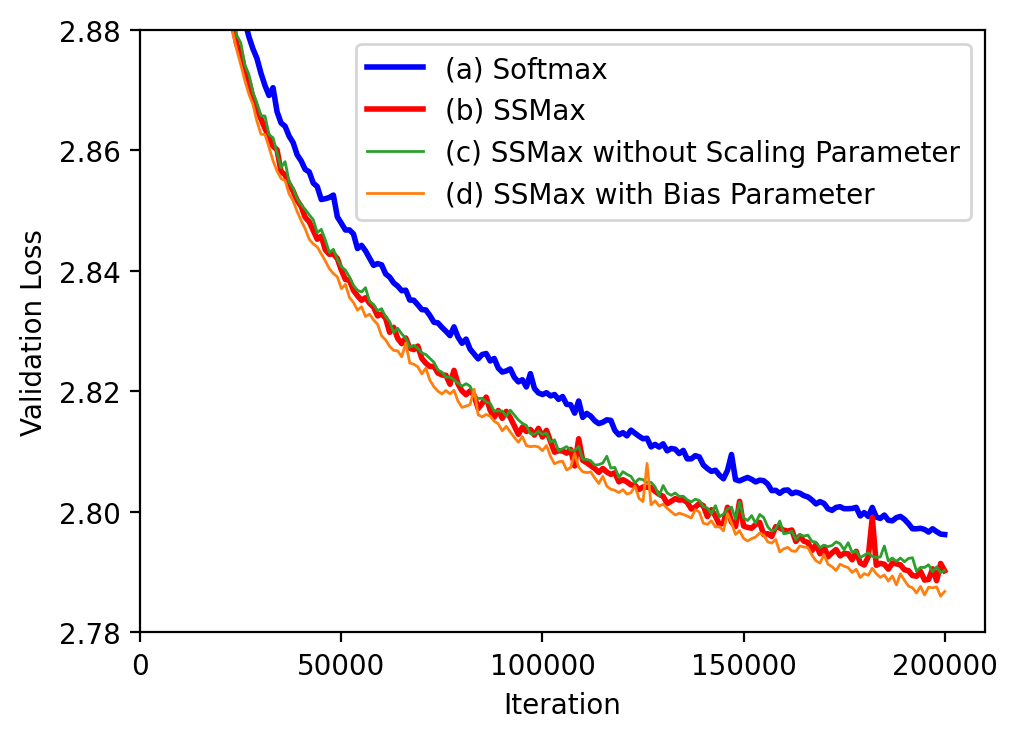}
    \vspace*{-1ex}
    \caption{
        Learning curves comparing the standard Transformer (a) and SSMax variants (b)--(d). All SSMax variants achieve consistently lower training loss compared to (a). Among them, the model with SSMax incorporating a bias parameter (d) exhibits the lowest loss throughout training. The results also indicate that removing the scaling parameter, as in (c), has little impact on the learning curve compared to (b).
    }
    \label{fig:learning_curve}
\end{figure}

We pretrained the models on the SlimPajama dataset~\cite{cerebras2023slimpajama}, a compressed version of RedPajama~\cite{together2023redpajama}. The models were trained with a batch size of 2048 using the AdamW optimizer~\cite{loshchilov2017decoupled}, with a learning rate of $6 \times 10^{-4}$. Training spanned 200,000 iterations, corresponding to approximately 419B tokens. Details of the pretraining hyperparameters are provided in \cref{apx:pt_param}.

We compared the learning curves of the standard Transformer (a) and SSMax variants (b)--(d). \Cref{fig:learning_curve} shows that all SSMax variants achieved consistently lower training loss compared to the standard Transformer (a). For example, SSMax (b) resulted in an approximate 0.008 reduction in training loss. Among the variants, the model with SSMax incorporating a bias parameter (d) exhibited the lowest loss throughout training. Additionally, a comparison between (b) and (c) indicates that removing the scaling parameter has little impact on the learning curve.

\subsection{Generalization to Longer Contexts}\label{sec:posloss}

\begin{figure}[t]
    \includegraphics[width=0.95\columnwidth]{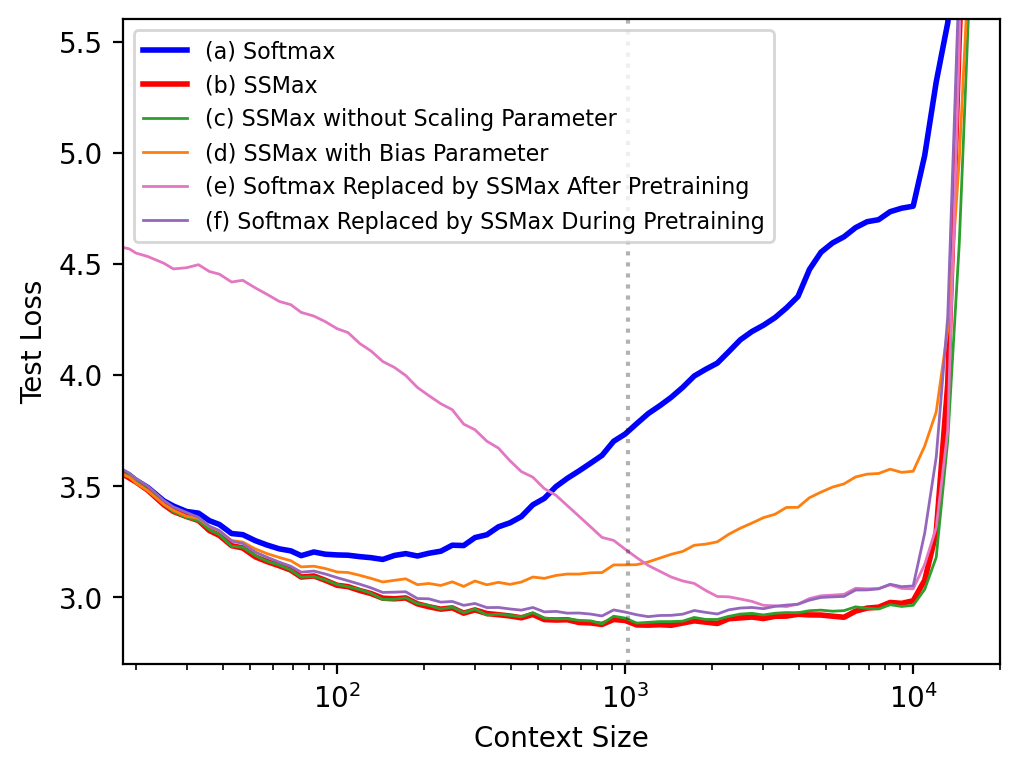}
    \vspace*{-1ex}
    \caption{
        Per-position test loss across context sizes up to 20,000. The x-axis represents context size, and the y-axis represents test loss. RoPE's $\theta$ was set to 50 times the training value, with no additional training after modification. The gray dotted line indicates the training sequence length of 1024. Results correspond to configurations (a)--(f). SSMax models (b) and (c) demonstrate improved long-context generalization compared to (a), while (d) exhibits degraded performance due to the bias parameter. Model (e), where Softmax was replaced with SSMax post-training, struggles with shorter contexts, whereas (f), which switched to SSMax during the final phase of pretraining, achieves performance somewhat close to (b), though not entirely equivalent.
    }
    \label{fig:posloss}
\end{figure}

To assess long-sequence modeling capabilities, we increased RoPE's $\theta$ from 10,000 to 500,000. Note that no additional training was performed after modifying $\theta$. This was done to isolate the impact of adjusting RoPE's $\theta$ and evaluate how well models generalize to longer contexts without further training. We evaluated models (a)--(f) by computing the per-position test loss using 100,000 sequences of length 20,000 randomly sampled from the SlimPajama test set. \Cref{fig:posloss} presents the results.

The standard Transformer (a) struggles significantly with extended context sizes, showing a noticeable increase in loss. Furthermore, (a) is not robust to substantial changes in RoPE's $\theta$, exhibiting degraded performance even at shorter context sizes compared to the original training setup. In contrast, SSMax models (b) and (c) maintain lower test loss across long contexts, demonstrating improved generalization to sequence lengths up to approximately 10 times the training sequence length. Additionally, \Cref{fig:posloss} indicates that SSMax models are notably more robust to modifications in RoPE's $\theta$ than (a).

The bias parameter (d), while improving training efficiency (\cref{sec:learning_curve}), weakens long-context performance. Although (d) still outperforms (a), it fails to fully preserve the benefits of SSMax, resulting in an intermediate performance between (a) and (b), indicating that the bias parameter degrades its long-context performance.

Regarding models where Softmax was replaced with SSMax after or during pretraining, \Cref{fig:posloss} shows that (e), in which Softmax was replaced with SSMax post-training, exhibits higher loss at shorter context sizes, suggesting that post-training replacement struggles to adapt to shorter contexts. Meanwhile, (f), where SSMax was introduced during the final phase of pretraining, achieves performance somewhat close to (b), though not entirely equivalent.

\subsection{Key Information Retrieval}\label{sec:niah}

\begin{figure}[!t]
    \centering
    \includegraphics[width=\columnwidth]{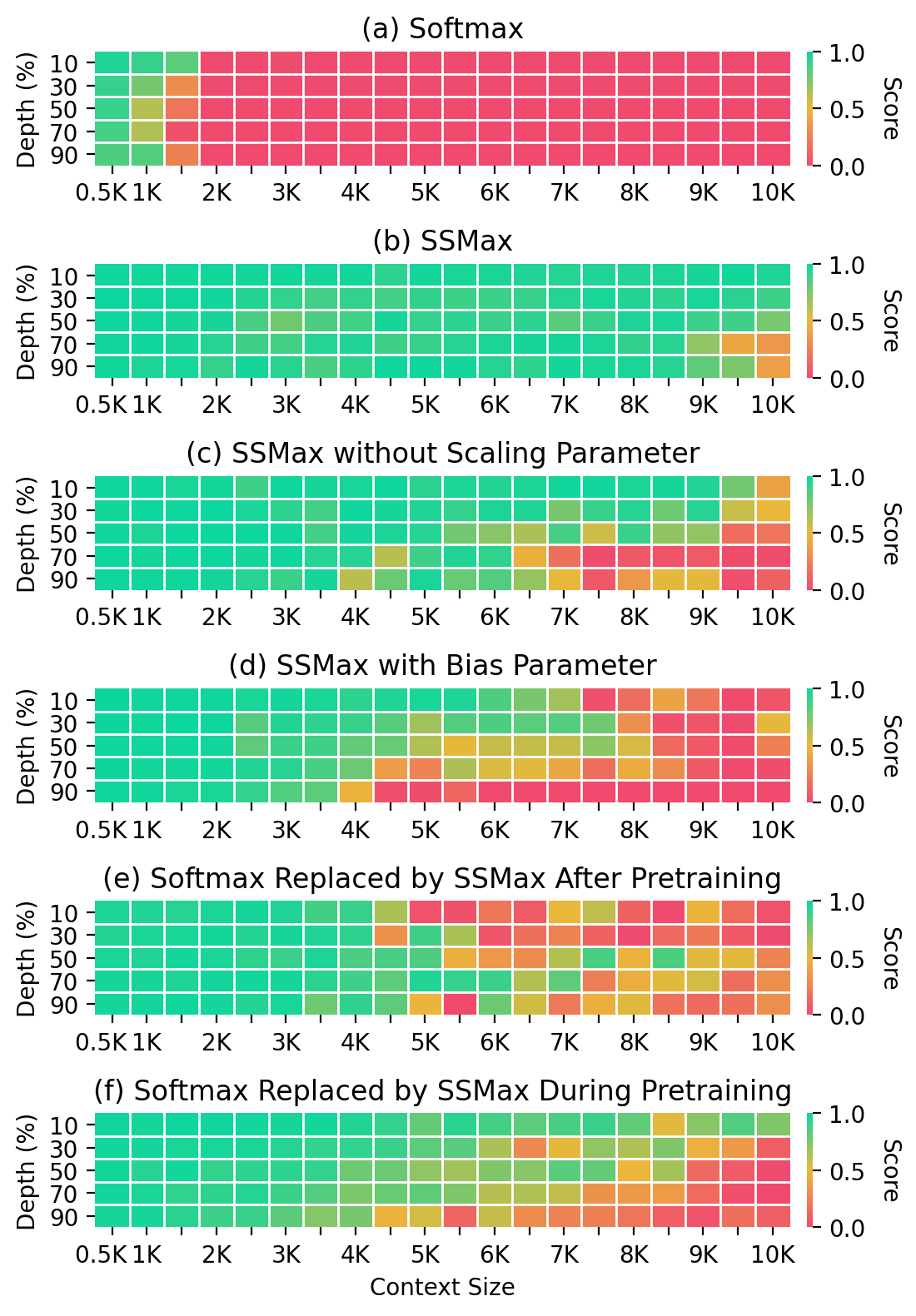}
    \vspace*{-3ex}
    \caption{
        Needle-In-A-Haystack test results. The horizontal axis represents context size, while the vertical axis denotes the depth at which the \textit{needle} is embedded within the context. Colors indicate retrieval accuracy. RoPE's $\theta$ was set to 500,000, a 50-fold increase from the pretraining value. The standard Transformer (a) fails to retrieve key information beyond short context sizes, while the SSMax model (b) maintains high retrieval accuracy even at context sizes approximately 10 times longer than in training. Models (c) and (d) show lower retrieval accuracy than (b), demonstrating that removing the scaling parameter or introducing a bias parameter degrades retrieval performance. Models where Softmax was replaced with SSMax after pretraining (e) and during pretraining (f) show partial improvements over (a) but remain far below (b).
    }
    \label{fig:niah}
\end{figure}

The Needle-In-A-Haystack test~\cite{kamradt2023niah} is widely used to evaluate a model's ability to retrieve key information embedded in a long context. In this study, we conduct a similar test, where the model must recall a randomly assigned seven-digit number corresponding to a randomly chosen city name, following the setup proposed in~\cite{arize2023niah}. Each sample contains a sentence, referred to as the \textit{needle}, such as \textit{``The special magic Tokyo number is: 8106422.''} This needle is inserted at a random location within the context, and the model is prompted to retrieve the correct number given the city name.

To evaluate key information retrieval, we first fine-tuned all six pretrained models (a)--(f) using Supervised Fine-Tuning. We used the SQuAD 2.0 dataset~\cite{rajpurkar2018know,rajpurkar2016squad}, selecting 86,820 examples where the context, question, and answer fields were all non-empty. Fine-tuning was conducted for 10 epochs, and the loss was computed as the sum of token-level losses for the answer span. Further details on fine-tuning hyperparameters are provided in \cref{apx:sft_param}.

Following fine-tuning, we increased RoPE's $\theta$ to 500,000, a 50-fold increase from training. No additional training was performed after modifying $\theta$, allowing us to assess generalization to longer contexts without adaptation.

For evaluation, we inserted the needle at five different depths within the context: 10\%, 30\%, 50\%, 70\%, and 90\%. Each configuration was tested with 1000 samples for each depth-context size combination. Decoding was performed greedily, selecting the most probable non-eos token at each step without sampling.

\Cref{fig:niah} presents the results of the Needle-In-A-Haystack test, demonstrating how different attention mechanisms affect key information retrieval performance in extended contexts. The standard Transformer (a) fails to retrieve key information beyond short context sizes. In contrast, the SSMax model (b) achieves significantly improved retrieval performance, successfully retrieving key information even at context sizes approximately 10 times longer than in training. Model (c), which removes the scaling parameter, performs noticeably worse than (b), despite demonstrating comparable performance in \cref{sec:learning_curve,sec:posloss}. This suggests that the scaling parameter plays a role in key information retrieval. Model (d), incorporating a bias parameter, performs better than (a) but exhibits a substantial degradation in performance compared to (b), indicating that the bias parameter negatively impacts key information retrieval.

For models where Softmax was replaced with SSMax after pretraining (e) or during pretraining (f), (e) achieves a substantial improvement over (a) but still performs noticeably worse than (b), suggesting limited adaptability when switching post-training. While (f) slightly outperforms (e), it remains significantly behind (b), highlighting the advantage of training with SSMax from the start rather than switching later.

These findings confirm SSMax's effectiveness in long-context retrieval, particularly when applied throughout pretraining.

\subsection{Attention Allocation to Key Information}\label{sec:needle_score}

\begin{figure}[t]
    \includegraphics[width=0.95\columnwidth]{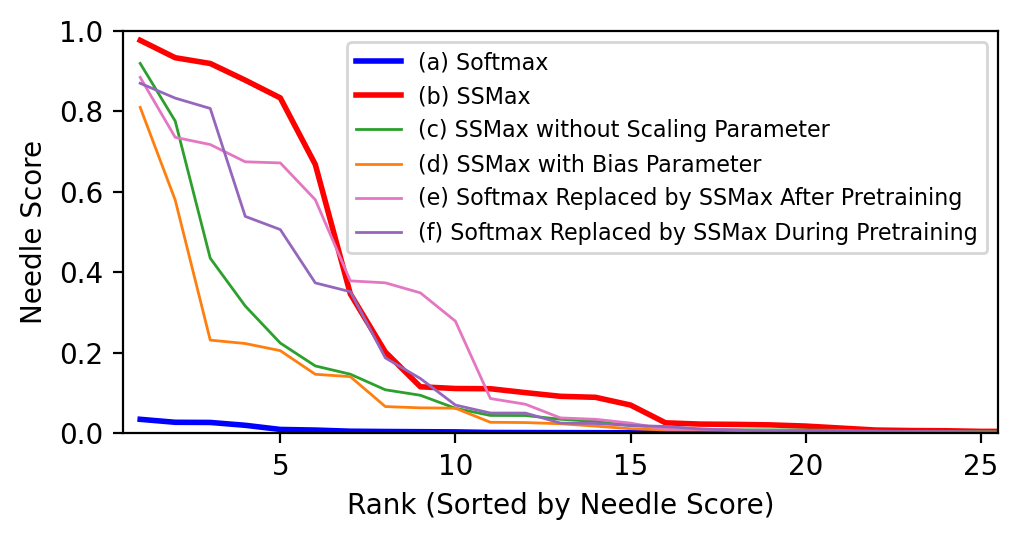}
    \vspace*{-1ex}
    \caption{
        Needle score distribution across attention layers and heads. The horizontal axis represents attention heads ranked by needle score (highest to lowest), while the vertical axis shows the corresponding needle score. Note that only the top 25 heads are shown for clarity, rather than all 144 heads. RoPE's $\theta$ was set to 500,000, a 50-fold increase from pretraining. The context size was 8000, with the needle sentence \textit{``The special magic Tokyo number is: 8106422.''} inserted at a depth of 50\%. The results demonstrate that the standard Transformer (a) fails to allocate significant attention to key tokens, whereas SSMax (b) effectively concentrates attention on them. Models (c), (d), (e), and (f) allocate more attention than (a) but fail to match the focus achieved by (b). Inference results indicate that (a) failed retrieval entirely, (b) and (c) successfully retrieved the correct number, and (d), (e), and (f) retrieved only the first digit but failed to recall the full number.
    }
    \label{fig:needle_score}
\end{figure}

\begin{figure*}[t]
    \centering
    \includegraphics[width=\textwidth]{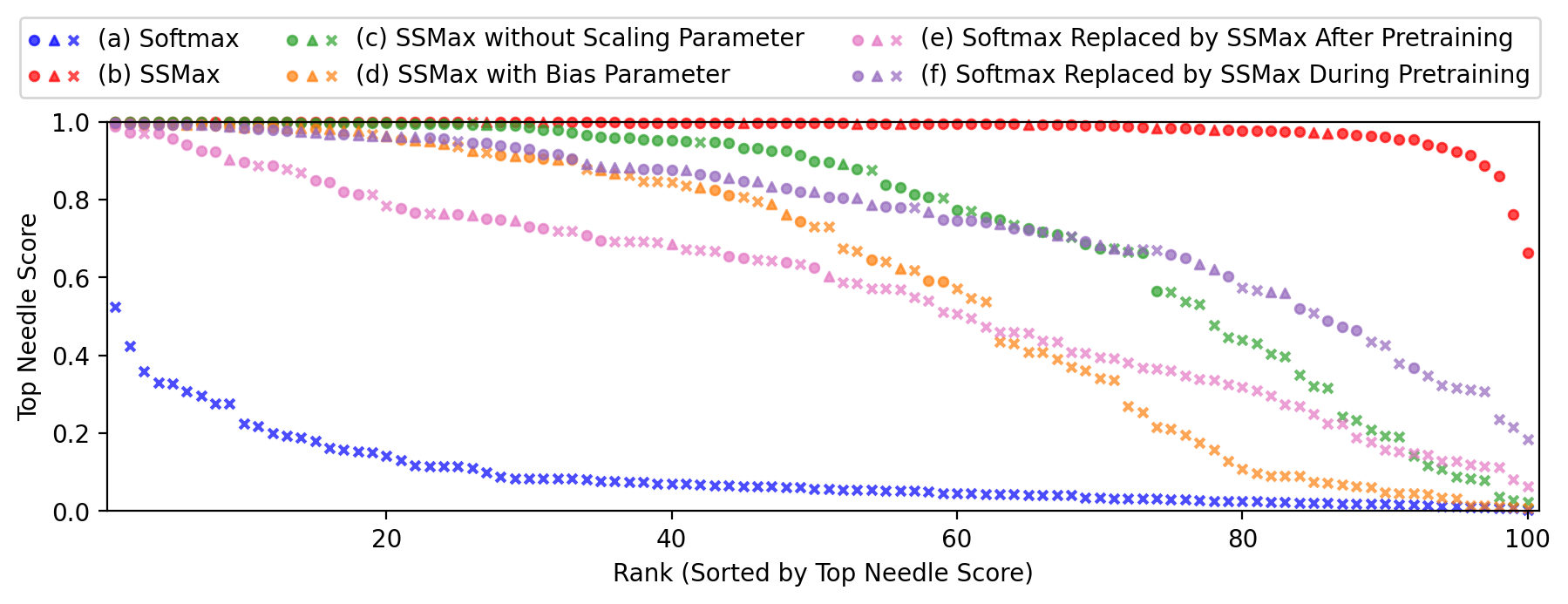}
    \vspace*{-4ex}
    \caption{
        Top needle score distribution across models. Each model was evaluated over 100 trials, and the highest needle score from each trial (corresponding to the leftmost value in \cref{fig:needle_score}) was recorded. The horizontal axis represents the rank of the top needle scores, sorted in descending order, while the vertical axis shows the corresponding score. Different markers indicate whether the retrieved number was fully correct ($\bullet$), incorrect but with the first digit correct ($\blacktriangle$), or completely incorrect ($\bm{\times}$). RoPE's $\theta$ was set to 500,000, a 50-fold increase from pretraining. Context size was fixed at 8000, with city names, numbers, and insertion depths randomly assigned. The results confirm that the standard Transformer (a) fails to focus attention on key tokens, whereas SSMax (b) exhibits strong concentration. Models (c), (d), (e), and (f) show partial improvements over (a) but fail to match (b)'s level of attention focus.
    }
    \label{fig:top_needle_score}
\end{figure*}

To further investigate key information retrieval, we analyzed how much attention each model allocates to key information during inference. Specifically, we define the \textit{needle score} as the sum of attention scores assigned to the span starting immediately after the colon (:) and extending to the end of the needle, fully covering the seven-digit number. This metric quantifies how effectively attention layers and heads focus on key information. The attention scores are measured when generating the first token of the model's response, which is expected to correspond to the initial part of the correct seven-digit number.

As a preliminary analysis, we examined needle scores in a fixed setting with RoPE's $\theta$ set to 500,000, a context size of 8000, and the needle sentence \textit{``The special magic Tokyo number is: 8106422.''} positioned at a depth of 50\%. \Cref{fig:needle_score} presents the needle scores for all layers and attention heads, sorted in descending order. This example highlights a stark contrast in attention allocation between models with SSMax and the standard Transformer. Among all models, model (b) demonstrates a particularly strong ability to focus on key information, achieving the highest needle scores. In contrast, the standard Transformer (a) fails to allocate meaningful attention to key information, with its highest needle scores remaining close to zero.

To obtain a more comprehensive understanding, we conducted a large-scale evaluation with 100 trials per model, keeping the context size fixed at 8000 tokens while varying the needle position and needle content. \Cref{fig:top_needle_score} presents the top needle scores observed in each trial, ranked in descending order. The figure also categorizes retrieval outcomes using different markers to indicate whether the full seven-digit number was correctly retrieved, only the first digit was correct, or the retrieval failed entirely. While not perfectly correlated, the results reveal a strong association between higher top needle scores and successful key information retrieval. Among all models, model (b) with SSMax consistently produces top needle scores that remain among the highest across trials, demonstrating its strong ability to focus attention on key information, whereas the standard Transformer (a) fails to allocate meaningful attention.

Model (c), which removes the scaling parameter, exhibits lower top needle scores than (b), despite previously demonstrating comparable performance in \cref{sec:learning_curve,sec:posloss}. This suggests that the scaling parameter contributes to more effective attention allocation to key information. Model (d), which incorporates a bias parameter, also shows reduced top needle scores, performing better than (a) but significantly worse than (b). Similarly, models where Softmax was replaced with SSMax after or during pretraining show partial improvements over (a), but their top needle scores remain lower than those of model (b). In particular, model (f), where SSMax was introduced in the later stage of pretraining, achieves higher top needle scores than model (e), yet still falls short of (b).

These results indicate that SSMax significantly enhances attention allocation to key information, especially when incorporated from the beginning of pretraining. Furthermore, they highlight the importance of the scaling parameter in maintaining effective attention distribution and suggest that introducing a bias parameter weakens attention to key information.

\section{Conclusion}

In this paper, we proposed Scalable-Softmax (SSMax), a novel alternative to Softmax in Transformer attention layers. SSMax addresses the issue of attention fading and enhances length generalization, enabling models to maintain attention over long contexts. Unlike Softmax, which suppresses attention scores as the input size increases, SSMax helps models retain focus on key information.

Through extensive evaluations, we demonstrated the effectiveness of SSMax across multiple aspects of Transformer performance. Models with SSMax consistently achieved lower loss values during pretraining, indicating improved optimization efficiency. When applied to longer contexts, these models retained significantly lower test loss than the standard Transformer, even as the context size extended well beyond the training sequence length. In key information retrieval tasks, SSMax models exhibited superior accuracy, successfully extracting relevant information even at context sizes up to ten times longer than those seen during training. Attention score analysis confirmed that SSMax improves attention allocation to key tokens, enhancing models' ability to retrieve key information in long-context scenarios.

While models trained with SSMax from the beginning of pretraining demonstrated the strongest generalization ability, we also found that models could benefit from SSMax even when introduced at later stages. Replacing Softmax with SSMax during or after pretraining led to noticeable improvements, demonstrating its adaptability as an enhancement for both newly trained and existing pretrained models.

These findings suggest that SSMax is a promising approach for addressing the limitations of Softmax in Transformer attention mechanisms, particularly for tasks involving extended contexts. In the future, SSMax has the potential to replace Softmax in the attention layers of all Transformer-based LLMs, including existing pretrained models. Its adaptability and ability to improve length generalization position it as a strong candidate for standard adoption in Transformer architectures.

\section*{Acknowledgements}

This research was conducted using NVIDIA GPGPU at the Center of Innovations for Sustainable Quantum AI (JST Grant Number JPMJPF2221), and the FUJITSU Supercomputer PRIMEHPC FX1000 and FUJITSU Server PRIMERGY GX2570 (Wisteria/BDEC-01) at the Information Technology Center, The University of Tokyo.
KMN is supported by the Daikin Endowed Research Unit: ``Research on Physics of Intelligence'', School of Science, the University of Tokyo, and the Center of Innovation for Sustainable Quantum AI (JST Grant Number JPMJPF2221).

\bibliography{main}

\begin{thebibliography}{32}
\providecommand{\natexlab}[1]{#1}
\providecommand{\url}[1]{\texttt{#1}}
\expandafter\ifx\csname urlstyle\endcsname\relax
  \providecommand{\doi}[1]{doi: #1}\else
  \providecommand{\doi}{doi: \begingroup \urlstyle{rm}\Url}\fi

\bibitem[Ainslie et~al.(2020)Ainslie, Ontanon, Alberti, Cvicek, Fisher, Pham, Ravula, Sanghai, Wang, and Yang]{ainslie2020etc}
Ainslie, J., Ontanon, S., Alberti, C., Cvicek, V., Fisher, Z., Pham, P., Ravula, A., Sanghai, S., Wang, Q., and Yang, L.
\newblock Etc: Encoding long and structured inputs in transformers.
\newblock \emph{Proceedings of the 2020 Conference on Empirical Methods in Natural Language Processing (EMNLP)}, pp.\  268--284, 2020.

\bibitem[{Arize AI}(2023)]{arize2023niah}
{Arize AI}.
\newblock Needle in a haystack - pressure testing llms, 2023.
\newblock URL \url{https://github.com/Arize-ai/LLMTest_NeedleInAHaystack2}.
\newblock Accessed on Jan 19, 2024.

\bibitem[Beltagy et~al.(2020)Beltagy, Peters, and Cohan]{beltagy2020longformer}
Beltagy, I., Peters, M.~E., and Cohan, A.
\newblock Longformer: The long-document transformer.
\newblock \emph{arXiv preprint arXiv:2004.05150}, 2020.

\bibitem[Brown et~al.(2020)Brown, Mann, Ryder, Subbiah, Kaplan, Dhariwal, Neelakantan, Shyam, Sastry, Askell, Agarwal, Herbert-Voss, Krueger, Henighan, Child, Ramesh, Ziegler, Wu, Winter, Hesse, Chen, Sigler, Litwin, Gray, Chess, Clark, Berner, McCandlish, Radford, Sutskever, and Amodei]{brown2020language}
Brown, T., Mann, B., Ryder, N., Subbiah, M., Kaplan, J.~D., Dhariwal, P., Neelakantan, A., Shyam, P., Sastry, G., Askell, A., Agarwal, S., Herbert-Voss, A., Krueger, G., Henighan, T., Child, R., Ramesh, A., Ziegler, D., Wu, J., Winter, C., Hesse, C., Chen, M., Sigler, E., Litwin, M., Gray, S., Chess, B., Clark, J., Berner, C., McCandlish, S., Radford, A., Sutskever, I., and Amodei, D.
\newblock Language models are few-shot learners.
\newblock \emph{Advances in Neural Information Processing Systems}, 33:\penalty0 1877--1901, 2020.

\bibitem[Child et~al.(2019)Child, Gray, Radford, and Sutskever]{child2019generating}
Child, R., Gray, S., Radford, A., and Sutskever, I.
\newblock Generating long sequences with sparse transformers.
\newblock \emph{arXiv preprint arXiv:1904.10509}, 2019.

\bibitem[Computer(2023)]{together2023redpajama}
Computer, T.
\newblock Redpajama: An open source recipe to reproduce llama training dataset, April 2023.
\newblock URL \url{https://github.com/togethercomputer/RedPajama-Data}.

\bibitem[Gupta \& Berant(2020)Gupta and Berant]{gupta2020gmat}
Gupta, A. and Berant, J.
\newblock Gmat: Global memory augmentation for transformers.
\newblock \emph{arXiv preprint arXiv:2006.03274}, 2020.

\bibitem[Kamradt(2023)]{kamradt2023niah}
Kamradt, G.
\newblock Needle in a haystack - pressure testing llms, 2023.
\newblock URL \url{https://github.com/gkamradt/LLMTest_NeedleInAHaystack}.
\newblock Accessed on Jan 19, 2024.

\bibitem[Kazemnejad et~al.(2023)Kazemnejad, Padhi, Natesan~Ramamurthy, Das, and Reddy]{kazemnejad2023impact}
Kazemnejad, A., Padhi, I., Natesan~Ramamurthy, K., Das, P., and Reddy, S.
\newblock The impact of positional encoding on length generalization in transformers.
\newblock \emph{Advances in Neural Information Processing Systems}, 36:\penalty0 24892--24928, 2023.

\bibitem[Kitaev et~al.(2020)Kitaev, Kaiser, and Levskaya]{kitaev2020reformer}
Kitaev, N., Kaiser, {\L}., and Levskaya, A.
\newblock Reformer: The efficient transformer.
\newblock \emph{arXiv preprint arXiv:2001.04451}, 2020.

\bibitem[LeCun et~al.(1998)LeCun, Bottou, Bengio, and Haffner]{lecun1998gradient}
LeCun, Y., Bottou, L., Bengio, Y., and Haffner, P.
\newblock Gradient-based learning applied to document recognition.
\newblock \emph{Proceedings of the IEEE}, 86\penalty0 (11):\penalty0 2278--2324, 1998.

\bibitem[Liu et~al.(2024{\natexlab{a}})Liu, Yan, Zaharia, and Abbeel]{liu2024world}
Liu, H., Yan, W., Zaharia, M., and Abbeel, P.
\newblock World model on million-length video and language with blockwise ringattention.
\newblock \emph{arXiv preprint arXiv:2402.08268}, 2024{\natexlab{a}}.

\bibitem[Liu et~al.(2024{\natexlab{b}})Liu, Yan, An, Qiu, and Lin]{liu2024scaling}
Liu, X., Yan, H., An, C., Qiu, X., and Lin, D.
\newblock Scaling laws of ro{PE}-based extrapolation.
\newblock \emph{The Twelfth International Conference on Learning Representations}, 2024{\natexlab{b}}.

\bibitem[Loshchilov \& Hutter(2019)Loshchilov and Hutter]{loshchilov2017decoupled}
Loshchilov, I. and Hutter, F.
\newblock Decoupled weight decay regularization.
\newblock \emph{International Conference on Learning Representations}, 2019.

\bibitem[Press et~al.(2021)Press, Smith, and Lewis]{press2021train}
Press, O., Smith, N.~A., and Lewis, M.
\newblock Train short, test long: Attention with linear biases enables input length extrapolation.
\newblock \emph{arXiv preprint arXiv:2108.12409}, 2021.

\bibitem[Radford et~al.(2019)Radford, Wu, Child, Luan, Amodei, and Sutskever]{radford2019language}
Radford, A., Wu, J., Child, R., Luan, D., Amodei, D., and Sutskever, I.
\newblock Language models are unsupervised multitask learners.
\newblock \emph{OpenAI blog}, 2019.

\bibitem[Rajpurkar et~al.(2016)Rajpurkar, Zhang, Lopyrev, and Liang]{rajpurkar2016squad}
Rajpurkar, P., Zhang, J., Lopyrev, K., and Liang, P.
\newblock {SQ}u{AD}: 100,000+ questions for machine comprehension of text.
\newblock \emph{Proceedings of the 2016 Conference on Empirical Methods in Natural Language Processing}, pp.\  2383--2392, 2016.

\bibitem[Rajpurkar et~al.(2018)Rajpurkar, Jia, and Liang]{rajpurkar2018know}
Rajpurkar, P., Jia, R., and Liang, P.
\newblock Know what you don't know: Unanswerable questions for {SQ}u{AD}.
\newblock \emph{Proceedings of the 56th Annual Meeting of the Association for Computational Linguistics}, 2:\penalty0 784--789, 2018.

\bibitem[Ramachandran et~al.(2017)Ramachandran, Zoph, and Le]{ramachandran2017searching}
Ramachandran, P., Zoph, B., and Le, Q.~V.
\newblock Searching for activation functions.
\newblock \emph{arXiv preprint arXiv:1710.05941}, 2017.

\bibitem[Roy et~al.(2021)Roy, Saffar, Vaswani, and Grangier]{roy2021efficient}
Roy, A., Saffar, M., Vaswani, A., and Grangier, D.
\newblock Efficient content-based sparse attention with routing transformers.
\newblock \emph{Transactions of the Association for Computational Linguistics}, 9:\penalty0 53--68, 2021.

\bibitem[Shaw et~al.(2018)Shaw, Uszkoreit, and Vaswani]{shaw2018self}
Shaw, P., Uszkoreit, J., and Vaswani, A.
\newblock Self-attention with relative position representations.
\newblock \emph{North American Chapter of the Association for Computational Linguistics}, pp.\  464--468, 2018.

\bibitem[Shazeer(2020)]{shazeer2020glu}
Shazeer, N.
\newblock Glu variants improve transformer.
\newblock \emph{arXiv preprint arXiv:2002.05202}, 2020.

\bibitem[Soboleva et~al.(2023)Soboleva, Al-Khateeb, Myers, Steeves, Hestness, and Dey]{cerebras2023slimpajama}
Soboleva, D., Al-Khateeb, F., Myers, R., Steeves, J.~R., Hestness, J., and Dey, N.
\newblock {SlimPajama: A 627B token cleaned and deduplicated version of RedPajama}.
\newblock \url{https://www.cerebras.net/blog/slimpajama-a-627b-token-cleaned-and-deduplicated-version-of-redpajama}, June 2023.
\newblock URL \url{https://huggingface.co/datasets/cerebras/SlimPajama-627B}.

\bibitem[Su et~al.(2024)Su, Ahmed, Lu, Pan, Bo, and Liu]{su2024roformer}
Su, J., Ahmed, M., Lu, Y., Pan, S., Bo, W., and Liu, Y.
\newblock Roformer: Enhanced transformer with rotary position embedding.
\newblock \emph{Neurocomputing}, 568:\penalty0 127063, 2024.

\bibitem[Sukhbaatar et~al.(2019)Sukhbaatar, Grave, Bojanowski, and Joulin]{sukhbaatar2019adaptive}
Sukhbaatar, S., Grave, E., Bojanowski, P., and Joulin, A.
\newblock Adaptive attention span in transformers.
\newblock \emph{Proceedings of the 57th Annual Meeting of the Association for Computational Linguistics}, pp.\  331--335, 2019.

\bibitem[Touvron et~al.(2023)Touvron, Martin, Stone, Albert, Almahairi, Babaei, Bashlykov, Batra, Bhargava, Bhosale, Bikel, Blecher, Cant{\'o}n~Ferrer, Chen, Cucurull, Esiobu, Fernandes, Fu, Fu, Fuller, Gao, Goswami, Goyal, Hartshorn, Hosseini, Hou, Inan, Kardas, Kerkez, Khabsa, Kloumann, Korenev, Koura, Lachaux, Lavril, Lee, Liskovich, Lu, Mao, Martinet, Mihaylov, Mishra, Molybog, Nie, Poulton, Reizenstein, Rungta, Saladi, Schelten, Silva, Smith, Subramanian, Tan, Tang, Taylor, Williams, Kuan, Xu, Yan, Zarov, Zhang, Fan, Hall, Kambadur, Narang, Rodriguez, Stojnic, Edunov, and Scialom]{touvron2023llama}
Touvron, H., Martin, L., Stone, K.~R., Albert, P., Almahairi, A., Babaei, Y., Bashlykov, N., Batra, S., Bhargava, P., Bhosale, S., Bikel, D.~M., Blecher, L., Cant{\'o}n~Ferrer, C., Chen, M., Cucurull, G., Esiobu, D., Fernandes, J., Fu, J., Fu, W., Fuller, B., Gao, C., Goswami, V., Goyal, N., Hartshorn, A.~S., Hosseini, S., Hou, R., Inan, H., Kardas, M., Kerkez, V., Khabsa, M., Kloumann, I.~M., Korenev, A.~V., Koura, P.~S., Lachaux, M.-A., Lavril, T., Lee, J., Liskovich, D., Lu, Y., Mao, Y., Martinet, X., Mihaylov, T., Mishra, P., Molybog, I., Nie, Y., Poulton, A., Reizenstein, J., Rungta, R., Saladi, K., Schelten, A., Silva, R., Smith, E.~M., Subramanian, R., Tan, X., Tang, B., Taylor, R., Williams, A., Kuan, J.~X., Xu, P., Yan, Z., Zarov, I., Zhang, Y., Fan, A., Hall, M., Kambadur, M., Narang, S., Rodriguez, A., Stojnic, R., Edunov, S., and Scialom, T.
\newblock Llama 2: Open foundation and fine-tuned chat models.
\newblock \emph{arXiv preprint arXiv:2307.09288}, 2023.

\bibitem[Vaswani et~al.(2017)Vaswani, Shazeer, Parmar, Uszkoreit, Jones, Gomez, Kaiser, and Polosukhin]{vaswani2017attention}
Vaswani, A., Shazeer, N., Parmar, N., Uszkoreit, J., Jones, L., Gomez, A.~N., Kaiser, {\L}., and Polosukhin, I.
\newblock Attention is all you need.
\newblock \emph{Advances in Neural Information Processing Systems}, 30, 2017.

\bibitem[Wang et~al.(2024)Wang, Ji, Wu, Yan, Gui, Zhang, Huang, and Wang]{wang2024length}
Wang, J., Ji, T., Wu, Y., Yan, H., Gui, T., Zhang, Q., Huang, X., and Wang, X.
\newblock Length generalization of causal transformers without position encoding.
\newblock \emph{arXiv preprint arXiv:2404.12224}, 2024.

\bibitem[Wei et~al.(2022)Wei, Wang, Schuurmans, Bosma, ichter, Xia, Chi, Le, and Zhou]{wei2022chain}
Wei, J., Wang, X., Schuurmans, D., Bosma, M., ichter, b., Xia, F., Chi, E., Le, Q.~V., and Zhou, D.
\newblock Chain-of-thought prompting elicits reasoning in large language models.
\newblock \emph{Advances in Neural Information Processing Systems}, 35:\penalty0 24824--24837, 2022.

\bibitem[Ye et~al.(2024)Ye, Dong, Xia, Sun, Zhu, Huang, and Wei]{ye2024differential}
Ye, T., Dong, L., Xia, Y., Sun, Y., Zhu, Y., Huang, G., and Wei, F.
\newblock Differential transformer.
\newblock \emph{arXiv preprint arXiv:2410.05258}, 2024.

\bibitem[Zaheer et~al.(2020)Zaheer, Guruganesh, Dubey, Ainslie, Alberti, Ontanon, Pham, Ravula, Wang, Yang, and Ahmed]{zaheer2020big}
Zaheer, M., Guruganesh, G., Dubey, K.~A., Ainslie, J., Alberti, C., Ontanon, S., Pham, P., Ravula, A., Wang, Q., Yang, L., and Ahmed, A.
\newblock Big bird: Transformers for longer sequences.
\newblock \emph{Advances in Neural Information Processing Systems}, 33:\penalty0 17283--17297, 2020.

\bibitem[Zhang \& Sennrich(2019)Zhang and Sennrich]{zhang2019root}
Zhang, B. and Sennrich, R.
\newblock Root mean square layer normalization.
\newblock \emph{Advances in Neural Information Processing Systems}, 32, 2019.

\end{thebibliography}
\bibliographystyle{icml2025}

\newpage
\appendix
\onecolumn
\section{Model Architecture Details}\label{apx:model_param}

\Cref{tb:model_param} presents the architectural details of the models used in \cref{sec:explain_experimental,sec:evaluations}.

\begin{table}[h!]
\centering
\begin{tabular}{@{}ll@{}}
\toprule
\textbf{Parameter} & \textbf{Value} \\ \midrule
Number of Layers & 12 \\
Number of Attention Heads & 12 \\
Hidden Size & 768 \\
Feedforward Size & 2,048 \\
Vocabulary Size & 50,257 \\ \bottomrule
\end{tabular}
\caption{Model architecture details.}
\label{tb:model_param}
\end{table}

\section{Pretraining Hyperparameters}\label{apx:pt_param}

\Cref{tb:pt_param} presents the detailed hyperparameters for pretraining in \cref{sec:explain_experimental,sec:learning_curve}.

\begin{table}[h!]
\centering
\begin{tabular}{@{}ll@{}}
\toprule
\textbf{Parameter} & \textbf{Value} \\
\midrule
Optimizer & AdamW \\
Adam $\beta$ parameters & (0.9, 0.95) \\
Weight decay & 0.1 (applied only to parameters of rank $\geq 2$) \\
Gradient clipping threshold & 1.0 \\
Learning rate & $6\times 10^{-4}$ \\
Learning rate scheduler & Constant \\
Warmup steps & 1,000 \\
Sequence length & 1,024 \\
Batch size (tokens per update) & 2,048 (2,097,152 tokens) \\
RoPE $\theta$ & 10,000 \\
Dropout & 0.0 \\
Data type & bfloat16 \\
\bottomrule
\end{tabular}
\caption{Pretraining hyperparameters.}
\label{tb:pt_param}
\end{table}

\section{Supervised Fine-Tuning Hyperparameters}\label{apx:sft_param}

\Cref{tb:sft_param} presents the detailed hyperparameters for supervised fine-tuning in \cref{sec:niah}.

\begin{table}[h!]
\centering
\begin{tabular}{@{}ll@{}}
\toprule
\textbf{Parameter} & \textbf{Value} \\
\midrule
Optimizer & AdamW \\
Adam $\beta$ parameters & (0.9, 0.999) \\
Weight decay & 0.0 \\
Gradient clipping threshold & 1.0 \\
Learning rate & $2\times 10^{-5}$ \\
Learning rate scheduler & Cosine \\
Warmup period (epochs) & 1.0 \\
Sequence length & 1,024 \\
Batch size (tokens per update) & 128 (131,072 tokens) \\
RoPE $\theta$ & 10,000 \\
Dropout & 0.0 \\
Data type & bfloat16 \\
\bottomrule
\end{tabular}
\caption{Fine-tuning hyperparameters.}
\label{tb:sft_param}
\end{table}

\end{document}